\documentclass[letterpaper,10pt,conference]{ieeeconf}

\IEEEoverridecommandlockouts
\usepackage{graphicx}
\usepackage{amsmath,amssymb}
\usepackage{booktabs}
\usepackage[nocompress,nospace]{cite}
\usepackage{tabularx}
\usepackage{xcolor}
\usepackage{colortbl}
\usepackage[draft]{hyperref}
\usepackage{float}
\title{\LARGE \bf
ForceRFT: Refining VLA Actions through Force-Guided Residual Reinforcement Learning
}

\author{
Yichen Wang$^{1,6}$,
Chaoyang Zhang$^{6}$,
Xuqi Su$^{1,4}$,
Jun Ma$^{5}$,
Haiyue Zhu$^{3}$,
and Xiaocong Li$^{1,2,3,*}$%
\thanks{$^{1}$College of Information Science and Technology,
Eastern Institute of Technology, Ningbo,
Ningbo 315200, China.}%
\thanks{$^{2}$Zhejiang Key Laboratory of Industrial Intelligence
and Digital Twin, Eastern Institute of Technology, Ningbo,
Ningbo 315200, China.}%
\thanks{$^{3}$Department of Electrical and Computer Engineering,
National University of Singapore, Singapore 117583.}%
\thanks{$^{4}$School of Automation and Intelligent Sensing,
Shanghai Jiao Tong University,
Shanghai, China.}%
\thanks{$^{5}$Robotics and Autonomous Systems Thrust,
The Hong Kong University of Science and Technology (Guangzhou),
Guangzhou 511453, China.}%
\thanks{$^{6}$Jiangsu Key Laboratory of Advanced Food
Manufacturing Equipment and Technology,
Jiangnan University, Wuxi, Jiangsu, China.}%
\thanks{$^{*}$Corresponding author: xiaocongli@eitech.edu.cn.}%
}
\begin{document}

\maketitle
\thispagestyle{empty}
\pagestyle{empty}

\begin{abstract}
Force-conditioned vision-language-action (VLA) policies can respond to contact, but when trained solely on demonstrations, their recovery behavior may be limited by demonstration coverage, and they do not learn from deployment outcomes. Human corrective imitation provides additional recovery examples, but its objective matches local action targets without explicitly optimizing task return. We present ForceRFT, a force-guided residual reinforcement learning framework that learns contact-dependent corrections from human supervision and autonomous task outcomes. A frozen, demonstration-trained SmolVLA-based prior generates force-conditioned action chunks, while a lightweight residual actor refines individual end-effector pose commands using wrist feedback acquired during chunk execution. The decision-time wrist wrench, its temporal change, and the selected base motion condition both residual correction and value estimation. Human corrections supervise the residual actor, while verified autonomous transitions train the twin critics and support value-guided updates to the same actor. Bootstrapping is restricted to autonomous segments, preventing TD credit from crossing human-intervention boundaries. Real-robot experiments on plug insertion, ring-on-peg assembly, and whiteboard wiping show higher autonomous success rates than the evaluated demonstration-trained and residual-imitation baselines. Comparisons with residual imitation support value-guided residual optimization, while plug-insertion ablations indicate the benefit of direct execution-time wrist feedback.

\end{abstract}

\section{Introduction}
\label{sec:introduction}

Vision-language-action (VLA)
models~\cite{smolvla,pi0,pi05,lingbotvla2,g05}
have advanced robotic manipulation by connecting visual and
language understanding with action learning from robot
demonstrations~\cite{brohan2023rt2visionlanguageactionmodelstransfer,kim2024openvlaopensourcevisionlanguageactionmodel}.
However, vision alone provides only indirect information about
the physical interactions that arise in precision assembly and
other contact-rich manipulation tasks, particularly when contact
interfaces are partially or fully occluded.
Visually similar configurations can involve different contact
forces and moments, requiring different motion adjustments.
Wrist force--torque feedback therefore provides complementary
information for interpreting contact and guiding action selection.

Recent work has incorporated force information into robot policy
learning, from force-centric imitation and reactive
control~\cite{forcemimic,foar,forceflow,forcepolicy}
to force-aware VLA
systems~\cite{yu2026forcevla,fdvla,fmvla,facet0,fawam,favla,forcevla2,compliantvla}.
These methods incorporate physical feedback into policy inputs,
learning targets, or execution to make motion responsive to
contact conditions.
For policies trained solely on demonstrations, however, recovery
behavior is learned from a fixed dataset that may poorly cover
deviations encountered during deployment.
Fresh force measurements can guide action selection, but
subsequent successes and failures do not update the learned response.

Obtaining informative experience through autonomous exploration
can be costly in contact-rich manipulation.
Human-in-the-loop learning provides expert feedback on states
visited by the policy~\cite{dagger,hgdagger,thriftydagger},
and intervention-based methods prioritize corrective segments
within mixed human--robot experience~\cite{iwr,sirius}.
These corrections extend recovery experience beyond the initial
demonstrations, but imitation optimizes agreement with local
action targets without explicitly optimizing task return.
Reward-driven approaches complement corrective supervision with
task outcomes to improve the policy~\cite{hilserl,conrft,recap}.
For contact-rich manipulation, this motivates learning a
correction policy that uses wrist feedback to adjust the intended
motion and improves through autonomous task outcomes.

To learn these corrections, we introduce ForceRFT, a force-guided
residual reinforcement learning framework for contact-rich
manipulation.
A frozen, demonstration-trained VLA prior generates
force-conditioned action chunks, while a lightweight residual
actor refines individual end-effector pose commands using wrist
feedback acquired after chunk generation.
The residual actor and twin critics~\cite{td3} both condition on
the selected base motion, wrist wrench, and its temporal change.
Human corrections provide local supervision for the actor,
while autonomous transitions train the critics and support
value-guided refinement of the same residual policy.

We evaluate ForceRFT on three real-robot contact-rich manipulation
tasks: plug insertion, ring-on-peg assembly, and whiteboard wiping.
ForceRFT achieves higher autonomous success rates than
SmolVLA-SFT, Force Prior, and residual imitation across all three
tasks.
Force Prior executes the same force-conditioned motion prior
without residual correction.
The comparison with residual imitation examines the benefit of
value-guided optimization beyond matching human corrections,
while plug-insertion ablations assess the contribution of direct
execution-time wrist feedback.

Our contributions are threefold:
\begin{enumerate}
    \renewcommand{\labelenumi}{\arabic{enumi}.}

    \item We propose \textbf{ForceRFT}, a force-guided residual
    reinforcement learning framework for contact-rich manipulation.
    It combines a frozen, force-conditioned VLA prior with
    command-level residual corrections driven by wrist feedback
    during action-chunk execution.

    \item We formulate online residual learning in which both
    the residual actor and twin critics are conditioned on the
    selected base motion, wrist wrench, and its temporal change.
    Policy learning combines value-guided optimization from
    autonomous interaction with supervised imitation of human
    corrections.
    
    \item We demonstrate higher autonomous success rates than
    SmolVLA-SFT, Force Prior, and residual imitation on plug
    insertion, ring-on-peg assembly, and whiteboard wiping.
    Comparisons with residual imitation support the benefit of
    the value objective.
    Plug-insertion ablations further support the contributions
    of execution-time wrist feedback and its temporal change.
\end{enumerate}
\section{Related Work}
\label{sec:related_work}

\subsection{VLA for Contact-Rich Manipulation}
\label{sec:rw_contact_vla}

Force--torque feedback complements vision by revealing interaction
loads during contact-rich manipulation. Existing approaches
incorporate physical information into pretrained VLA models through
multimodal fusion, decoder conditioning, force-aware curricula,
and force-token distillation~\cite{yu2026forcevla,tavla,craft,fdvla}.
These strategies explore how measured wrist wrench, joint-torque
cues, or distilled force representations inform action generation.
Beyond instantaneous conditioning, temporal approaches encode
wrench histories and jointly predict actions and future wrenches
to retain contact events and anticipate interaction
dynamics~\cite{fmvla,facet0,fawam}.
Execution-time feedback further supports reactive motion through
force-conditioned action decoding at a higher rate than visual
inference and residual correction based on discrepancies between
predicted and measured wrenches~\cite{favla,fawam}.
A complementary direction connects task-conditioned predictions
to explicit contact regulation through hybrid force--position
actions and force-regulated variable impedance
control~\cite{forcevla2,compliantvla}.
Together, these approaches connect force representation,
contact prediction, and feedback control.
Building on force-aware multimodal fusion~\cite{yu2026forcevla},
ForceRFT combines force-conditioned motion generation with online
learning of corrections from wrist feedback during execution.

\subsection{Online Reinforcement Learning for VLA}
\label{sec:rw_online_rl}

Online reinforcement learning for VLA must translate sparse task
outcomes into policy improvement while reusing demonstrations and
experience collected under changing behavior policies.
ConRFT~\cite{conrft} incorporates Q-guided improvement into a
consistency-policy objective regularized by behavior cloning,
whereas RECAP~\cite{recap} uses estimated advantages to condition
generative policy training.
These formulations connect value estimation to policy improvement
through different mechanisms: action optimization in the former
and conditional policy extraction in the latter.
Other approaches restrict the scope of reward-driven updates.
iRe-VLA~\cite{irevla} alternates action-head reinforcement learning
with supervised consolidation of the complete model, while
ResiP~\cite{ankile2024imitationrefinementresidual}, Policy Decorator~\cite{policydecorator},
and ResFiT~\cite{ankile2025resfit}
formulate adaptation as learning residual corrections to an
existing policy, following the residual reinforcement learning
paradigm~\cite{johannink2018residualreinforcementlearningrobot}.
Human interventions provide an additional source of recovery
experience and corrective supervision, supporting off-policy
learning when autonomous exploration is costly~\cite{hilserl,conrft}.
Their use also raises a credit-assignment question, since success following human recovery does not by itself establish the quality of autonomous actions preceding intervention.
For contact-rich adaptation, TORL-VLA~\cite{zheng2026torlvla} uses
measured and predicted wrench cues to refine action references
while restricting credit across intervention boundaries, and
Facet-0~\cite{facet0} evaluates joint action--wrench proposals to
guide policy improvement.
In contrast, ForceRFT centers online learning on how measured
wrist feedback informs motion correction and its value during
execution. Task rewards and human correction targets train this
response without requiring future-wrench prediction.
\section{Preliminaries}
\label{sec:problem_formulation}

\subsection{Problem Formulation}
\label{sec:contact_rich_manipulation}

We study contact-rich robot manipulation with human intervention.
Given a language instruction $\ell$, the robot must generate motion
and gripper commands to complete the specified task.

At time $t$, the available observation is
\begin{equation}
o_t^F=
\left(I_t^{\mathrm{ext}},I_t^{\mathrm{wrist}},
\ell,\mathbf s_t,\mathbf w_t\right),
\label{eq:force_observation}
\end{equation}
where $I_t^{\mathrm{ext}}$ and $I_t^{\mathrm{wrist}}$ denote
external and wrist RGB images, respectively.
The proprioceptive input $\mathbf s_t\in\mathbb R^7$ represents
the measured tool center point (TCP) position, roll--pitch--yaw
orientation, and gripper width. The TCP is located midway between
the gripper fingertips. The wrist wrench $\mathbf w_t\in\mathbb R^6$
contains three force and three moment components, with moments
referenced at the TCP and both vectors expressed in the robot
base frame. State and wrench inputs are normalized using fixed
demonstration statistics.

Let $\bar{\mathbf s}$ denote the measured robot state in physical
coordinates, and let $\bar{\mathbf a}$ specify an absolute TCP
pose target and gripper opening.
The action representation relative to this measured state is
$\mathcal A_{\bar{\mathbf s}}(\bar{\mathbf a})
=\mathcal N_a(\bar{\mathbf a}\ominus\bar{\mathbf s})$.
The operator $\ominus$ subtracts base-frame TCP positions and
wraps componentwise RPY differences to $[-\pi,\pi)$ within the
controller's locally bounded orientation convention; the gripper
target remains absolute. The action normalizer $\mathcal N_a$
is fixed after demonstration training.
Controller acceptance records a command target; the resulting
TCP motion is observed through the robot state.

During online learning, an operator may take over and provide
corrective pairs $(o_t^F,\bar{\mathbf a}_t^{\mathrm H})$.
Episode outcomes from a frozen reward detector are verified
against operator annotations. A confirmed success receives
terminal reward $1$; all other rewards are zero.
Starting from demonstrations, our objective is to improve
autonomous task completion through task feedback and human
corrections.

\subsection{Motion Prior}
\label{sec:forcerft_architecture}

Let $i$ index base-policy inference requests and $n$ index command
decisions. The demonstration-trained prior $\pi_\theta$ maps
$o_i^F$ to a chunk
$\mathbf A_i^{\mathrm{base}}
=(\mathbf a_{i,0}^{\mathrm{base}},\ldots,
\mathbf a_{i,H-1}^{\mathrm{base}})$,
whose $H$ actions are normalized relative to the measured request
state $\bar{\mathbf s}_i^{\mathrm{req}}$.
We instantiate the prior with SmolVLA~\cite{smolvla}, incorporating
wrench--vision--language fusion following ForceVLA~\cite{yu2026forcevla}.
Section~\ref{sec:compact_motion_prior} describes its conditioning
and training. The complete prior remains fixed during online
adaptation.

\section{ForceRFT}
\label{sec:forcerft}

\subsection{Overview}
\label{sec:forcerft_overview}

\begin{figure*}[t]
    \centering
    \includegraphics[width=0.95\textwidth]{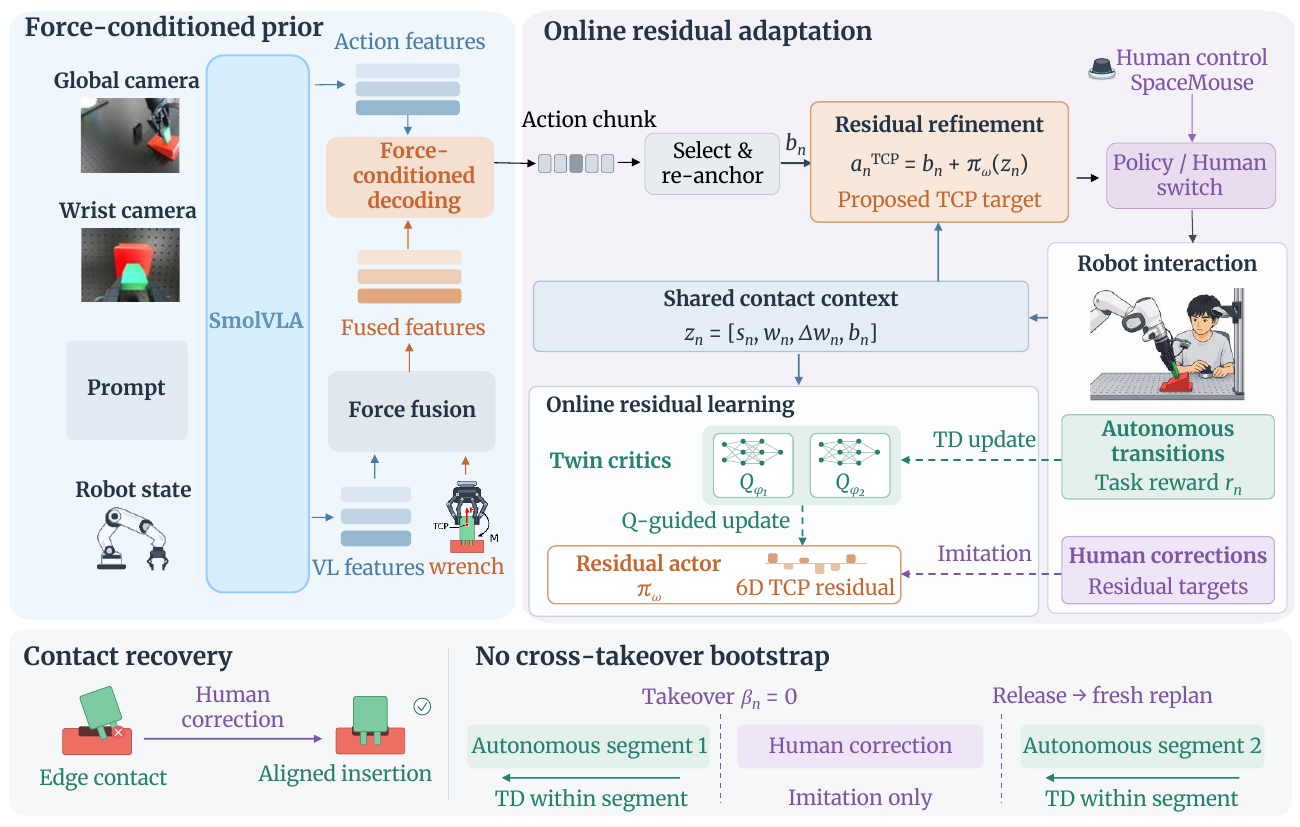}
    \caption{Overview of ForceRFT.
    A demonstration-trained force-conditioned prior generates
    action chunks and remains frozen online.
    At each execution decision, the selected TCP target is
    re-anchored to the current robot state and refined by a bounded
    residual conditioned on the state, base motion, and wrist feedback.
    Only autonomous transitions train the twin critics and support
    value-guided actor updates; human corrections supervise the
    same actor through residual imitation.
    Takeover transfers control to the operator and blocks
    cross-takeover bootstrapping ($\beta_n=0$); release triggers
    fresh replanning.
    Solid arrows denote control/data flow; dashed arrows denote
    learning updates.}
    \label{fig:forcerft_framework}
\end{figure*}

As illustrated in Fig.~\ref{fig:forcerft_framework}, ForceRFT
combines force-conditioned motion generation with execution-time
residual adaptation. A demonstration-trained prior generates
action chunks from the task context and wrist feedback.
During execution, a residual actor adjusts each selected TCP
target using the current robot state, wrist wrench, and
time-normalized wrench increment, incorporating contact changes
observed after chunk generation. The prior remains frozen online,
while the residual actor learns from value-guided updates on
autonomous experience and supervised imitation of human corrections.

\subsection{Force-Conditioned Decoding}
\label{sec:compact_motion_prior}

The appropriate response to a wrist load depends on the intended
motion and surrounding scene. We therefore condition action
decoding on the joint task and contact context.

Following ForceVLA~\cite{yu2026forcevla}, a wrench encoder $\phi_w$
and an attention-based fusion module with sparse mixture-of-experts
refinement produce $\mathbf Z_i^F=\mathcal F_{\mathrm{MoE}}
(\mathbf P_i,\phi_w(\mathbf w_i))$, where $\mathbf P_i$ contains
vision--language features. We provide this fused context to the
SmolVLA action expert through action-conditioned cross-attention:
\begin{equation}
\widetilde{\mathbf h}_{i,j}^{\tau}
=\mathbf h_{i,j}^{\tau}
+\tanh(\alpha)\mathbf W_o
\operatorname{Attn}(\mathbf q_{i,j}^{\tau},\mathbf Z_i^F).
\label{eq:force_conditioned_action_features}
\end{equation}
Here, $\mathbf h_{i,j}^{\tau}$ is the action feature for slot $j$
at flow time $\tau$, and $\alpha$ and $\mathbf W_o$ are a learned
gating parameter and output projection, respectively.
The query $\mathbf q_{i,j}^{\tau}$ combines the action feature
with the noisy action, flow-time encoding, and slot embedding.
Access to the fused contact information thus depends on the
action being decoded and can vary across slots and flow steps.
The gated update enters the action features before projection
to the velocity field $v_\theta$.

We train the prior on demonstrations using conditional flow
matching. Let $\mathbf A_i^{\mathrm{exp}}$ contain $H$ normalized
demonstration actions relative to the request pose.
For noise $\boldsymbol\epsilon_i\sim\mathcal N(\mathbf 0,\mathbf I)$
and flow time $\tau\in[0,1]$, the interpolated input is
$\mathbf x_i^\tau=(1-\tau)\mathbf A_i^{\mathrm{exp}}
+\tau\boldsymbol\epsilon_i$.
The training objective is
\begin{equation}
\begin{aligned}
\mathcal L_{\mathrm{SFT}}
&=\mathbb E\!\left[
\left\|v_\theta(\mathbf x_i^\tau,\tau;o_i^F)
-(\boldsymbol\epsilon_i-\mathbf A_i^{\mathrm{exp}})
\right\|_M^2\right]\\
&\quad+\mathcal L_{\mathrm{route}},
\end{aligned}
\label{eq:force_prior_training}
\end{equation}
where the expectation is over demonstrations, noise, and flow time.
The masked norm averages over valid action coordinates and slots,
and $\mathcal L_{\mathrm{route}}$ collects the weighted
expert-routing regularizers. After training, the complete prior
is frozen, and online adaptation operates on its execution commands.

\subsection{Residual Adaptation}
\label{sec:force_guided_residual}

Following residual policy refinement~\cite{ankile2024imitationrefinementresidual},
we adjust each selected base command using contact feedback
sampled at its execution decision. The mapping
$n\mapsto(i(n),j(n))$ associates decision $n$ with an active chunk
and its selected slot. At selection time $t_n^{\mathrm{sel}}$,
the slot is
$j(n)=\lceil f_m(t_n^{\mathrm{sel}}-t_{i(n)}^{\mathrm{ref}})\rceil$,
where $t_i^{\mathrm{ref}}$ is the request observation time and
$f_m=30\,\mathrm{Hz}$ is the model timebase.
The $H=50$ slots are indexed from zero; a chunk with $j(n)\geq H$
has expired and is replaced through replanning.

Each chunk permits $M$ command dispatches accepted by the
controller. When at most $L$ remain and no request is pending,
we request a new chunk asynchronously. A fresh result can replace
the active chunk before its dispatch budget is exhausted.
The online loop defaults to $M=8$ and $L=7$.
Takeover invalidates the active chunk and outstanding requests;
release initiates inference from a fresh observation.

We recover the selected absolute target from the request state
and re-express it relative to the measured state at decision $n$:
\begin{equation}
\mathbf b_n
=\mathbf P_{\mathrm{TCP}}\mathcal A_{\bar{\mathbf s}_n}\!\left[
\mathcal A_{\bar{\mathbf s}_{i(n)}^{\mathrm{req}}}^{-1}
(\mathbf a_{i(n),j(n)}^{\mathrm{base}})\right],
\label{eq:base_execution_mapping}
\end{equation}
where $\mathbf P_{\mathrm{TCP}}$ selects the six pose coordinates.
The residual actor receives
$\mathbf z_n=[\mathbf s_n^\top,\mathbf w_n^\top,
\Delta\mathbf w_n^\top,\mathbf b_n^\top]^\top$.
State and base motion contextualize the current wrist load and
its recent change. Unequal feedback intervals are accounted for by
\begin{equation}
\Delta\mathbf w_n
=\frac{\Delta t_{\mathrm{ref}}}{t_n-t_{n^-}}
(\mathbf w_n-\mathbf w_{n^-}).
\label{eq:wrench_change}
\end{equation}
Here, $t_n$ is the feedback sampling time for the residual decision,
$n^-$ is the preceding decision in the same control segment,
and $\Delta t_{\mathrm{ref}}=0.1\,\mathrm{s}$.
The increment is zero at the first decision of each segment.

The bounded actor $\pi_\omega(\mathbf z_n)
=c_{\mathrm{res}}\tanh f_\omega(\mathbf z_n)$ produces the correction,
where $c_{\mathrm{res}}>0$ bounds each normalized residual component.
The proposed TCP target in decision-relative normalized
coordinates is
\begin{equation}
\mathbf a_n^{\mathrm{TCP}}
=\mathbf b_n+\pi_\omega(\mathbf z_n).
\label{eq:forcerft_residual_overview}
\end{equation}
We convert this proposal to an absolute target using the same
measured decision state and inverse normalizer.
During autonomous execution, the gripper follows the base command.
Controller processing may modify the proposed target before
acceptance. The output head of the actor is initialized to zero,
so the initial autonomous proposals reproduce those of the prior.

\subsection{Human-in-the-Loop Reinforcement Learning}
\label{sec:interaction_and_human_learning}

ForceRFT learns residual corrections from task rewards and human
supervision~\cite{hilserl,conrft}.
The actor and critics share the state, wrist feedback, and selected
base motion as context. Autonomous transitions train the critics
and support value-guided actor updates, while human corrections
provide supervised targets for the same actor.

Let $\boldsymbol\delta_n^{\mathrm{prop}}$ denote the recorded
residual proposed before controller processing. Critic fitting,
value-guided actor optimization, and target-policy evaluation
all use this proposal representation. Controller processing forms
part of the environment dynamics, with successor observations
and rewards describing the resulting transition.
The critics additionally receive the base gripper target through
$\boldsymbol\xi_n=[\mathbf z_n^\top,b_n^g]^\top$,
where $b_n^g$ is the normalized base gripper command.
This distinguishes different gripper commands when the current
gripper width and TCP context coincide.
The compact context supports approximate value estimation under
partially observed contact and controller dynamics.

Replay $\mathcal R_{\mathrm{pol}}$ contains autonomous transitions
from successful or failed episodes admitted after record and
execution validation and operator--detector outcome agreement.
Records include proposals, decision observations, controller
acceptance, and verified segment boundaries or immediate successors.
Holding an existing command or rejecting a proposal without
execution does not create a new decision.
An autonomous segment ends at task termination, takeover, or reset;
human release starts a fresh autonomous segment.
We set $\beta_n=1$ for a verified immediate successor within the
same segment and $\beta_n=0$ at an actual segment boundary.
A nonboundary record without a verified immediate successor is
excluded from TD learning; it is neither treated as terminal nor
linked to a later decision.

Let $Q_{\min}$ be the smaller critic estimate; overbars denote
target networks. In each autonomous minibatch,
$\mathcal B_{\mathrm{TD}}$ retains boundary records and nonboundary
records whose composed target-policy commands at verified
successors pass recorded checks on finiteness, workspace,
orientation, gripper limits, and command continuity.
Filtering leaves historical TD eligibility and $\beta_n$ unchanged.
For nonempty $\mathcal B_{\mathrm{TD}}$, the TD target and critic
loss are
\begin{equation}
\begin{aligned}
y_n
&=r_n+\gamma\beta_n\bar Q_{\min}\!\left(
\boldsymbol\xi_{n+1},\pi_{\bar\omega}(\mathbf z_{n+1})\right),\\
\mathcal L_Q
&=\frac12\sum_{k=1}^{2}
\mathbb E_{\mathcal B_{\mathrm{TD}}}\!\left[
\left(Q_{\phi_k}(\boldsymbol\xi_n,\boldsymbol\delta_n^{\mathrm{prop}})
-y_n\right)^2\right].
\end{aligned}
\label{eq:force_residual_td}
\end{equation}
Here, $\gamma$ discounts each decision transition, and $r_n$ is
the sparse terminal reward defined in
Section~\ref{sec:contact_rich_manipulation}.
At an actual segment boundary, $y_n=r_n$ without evaluating a
successor. Terminal reward remains attached to the actual terminal
decision. This defines a segment-return surrogate for autonomous
task completion, restricting TD credit to the current autonomous
segment~\cite{zheng2026torlvla}. Success after intervention provides
no TD credit to earlier segments; human recovery motions instead
contribute through imitation.

During intervention, the pre-takeover absolute base target remains
fixed and is re-expressed at each human decision pose.
Let $\mathbf a_n^{\mathrm{H,acc}}$ denote the controller-accepted
human TCP target in the same decision-relative normalized
coordinates as $\mathbf b_n$.
The human residual target is $\boldsymbol\delta_n^{\mathrm H}
=\mathbf a_n^{\mathrm{H,acc}}-\mathbf b_n$.
For imitation, we project it onto the permitted residual range:
$\widetilde{\boldsymbol\delta}_n^{\mathrm H}
=\Pi_{\mathcal C}(\boldsymbol\delta_n^{\mathrm H})$,
where $\mathcal C=[-c_{\mathrm{res}},c_{\mathrm{res}}]^6$.
Projection changes only the supervision target, leaving the
accepted command and recorded outcome unchanged.
Human transitions are excluded from critic learning because the
fixed absolute base target and independently controlled gripper
produce different execution conditions from autonomous control.

The actor objective combines value improvement, human imitation,
and residual regularization:
\begin{equation}
\mathcal L_\pi
=\lambda_Q\mathcal L_V+\lambda_H\mathcal L_H+\lambda_R\mathcal L_R,
\label{eq:force_residual_actor_objective}
\end{equation}
where $\lambda_Q,\lambda_H,\lambda_R\geq0$ and
\begin{align*}
\mathcal L_V
&=-\mathbb E_{\mathcal R_{\mathrm{pol}}^Q}\!\left[
Q_{\min}(\boldsymbol\xi_n,\pi_\omega(\mathbf z_n))\right],\\
\mathcal L_H
&=\frac16\mathbb E_{\mathcal R_{\mathrm{human}}}\!\left[
\|\pi_\omega(\mathbf z_n)-\widetilde{\boldsymbol\delta}_n^{\mathrm H}\|_2^2
\right],\\
\mathcal L_R
&=\frac16\mathbb E_{\mathcal R_{\mathrm{reg}}}\!\left[
\|\pi_\omega(\mathbf z_n)\|_2^2\right].
\end{align*}
Here, $\mathcal R_{\mathrm{pol}}^Q$ supplies verified autonomous
contexts whose current actor proposals pass the same checks;
$\mathcal R_{\mathrm{human}}$ supplies corrections from successful
and failed episodes.
For $\mathcal R_{\mathrm{reg}}$, we use autonomous contexts when
available and human contexts otherwise.
With the actor initialized to zero residual and the critics
initialized with zero residual-input weights, human supervision
initiates adaptation. Subsequent autonomous executions provide
reward-based refinement. Only the residual actor and critics
are optimized online.

After a one-off critic-only warm-up, training permits at most
$\lfloor N_{\mathrm{TD}}/8 \rfloor$ completed joint cycles,
where $N_{\mathrm{TD}}$ counts unique autonomous transitions from
admitted episodes that satisfy the historical TD eligibility rules above.
Each cycle comprises two executed twin-critic optimizer updates
and one actor-update attempt. An empty $\mathcal B_{\mathrm{TD}}$
defers the remaining updates without completing the cycle.
Replay resampling and per-update target filtering do not change
$N_{\mathrm{TD}}$. Training pauses when the cycle allowance
is exhausted.

\section{EXPERIMENTS}
\label{sec:experiments}
We evaluate ForceRFT on three real-world contact-rich manipulation
tasks: plug insertion, ring-on-peg assembly, and whiteboard wiping.
Our experiments address four research questions:
(1) whether force-conditioned action generation improves autonomous
performance over a demonstration-trained policy without force input;
(2) whether value-based residual optimization provides additional
gains beyond imitating human corrections;
(3) how autonomous success evolves with online interaction; and
(4) how decision-time wrist wrench and its increment contribute
to residual adaptation.
We compare four methods across all three tasks and conduct
targeted wrist-feedback ablations on the insertion task.

\subsection{Experimental Setup}
\label{sec:experimental_setup}

\begin{figure}[t]
    \vspace*{1.5mm}
    \centering
    \includegraphics[width=0.9\columnwidth]{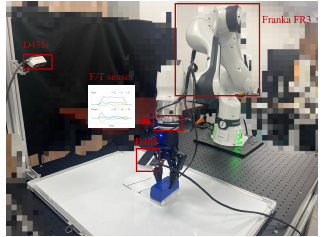}
\caption{\textbf{Experimental platform.} Real-robot setup with
dual-view RGB cameras and wrist force/torque sensing.}
    \label{fig:forcerft_setup}
\end{figure}

\begin{figure*}[t]
    \vspace*{1.5mm}
    \centering
    \includegraphics[width=0.9\textwidth]{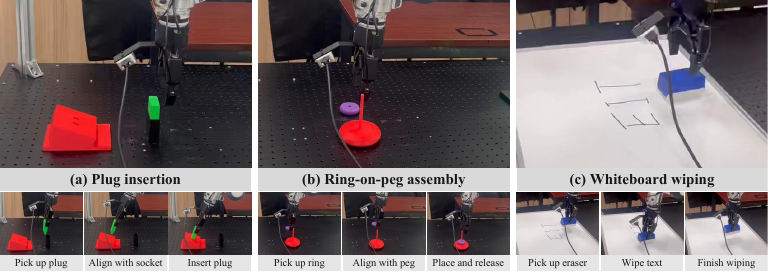}
    \caption{\textbf{Task visualization.}
    Representative stages of plug insertion,
    ring-on-peg assembly, and whiteboard wiping.}
    \label{fig:task_stages}
\end{figure*}

\noindent\textbf{Tasks and Platform.}
We use a Franka Research 3 with a Robotiq 2F-85 gripper,
an OnRobot HEX-E wrist force/torque sensor, and external
(D435i) and wrist (D405) RGB cameras. Human corrections
are provided through a SpaceMouse Compact
(Fig.~\ref{fig:forcerft_setup}).
Fig.~\ref{fig:task_stages} illustrates the three tasks.
Insertion and assembly require fully inserting the plug
or seating the ring on the peg, respectively, followed
by release. Wiping requires removing all marked text,
lifting the eraser off the board, and stopping.

\noindent\textbf{Evaluation Protocol.}
Each setting is evaluated over 30 autonomous rollouts
with fixed policy parameters and the same task-specific
initialization protocol. Success requires full task
completion within 60~s for insertion and assembly,
or 180~s for wiping, without human intervention,
within-trial resets, or protective stops.
We report successful trials out of 30.
The main comparison and checkpoint evaluations cover all
three tasks, while wrist-feedback ablations focus on
plug insertion.

\noindent\textbf{Baselines and Training.}
We compare SmolVLA-SFT without force input,
the force-conditioned prior without residual correction,
residual imitation, and ForceRFT.
Residual imitation sets $\lambda_Q=0$ while retaining
human corrective supervision and residual regularization.
All methods share the same 50 demonstrations per task.
Each online variant uses a 30-minute data-collection
and training budget per evaluated task, with the same
takeover criteria.
For ForceRFT, critic learning starts once at least
1000 unique TD-eligible autonomous transitions from
at least three admitted episodes are available.
A one-off critic-only warm-up of 256 twin-critic
optimizer updates precedes joint training, whose
data-dependent update allowance is defined in
Section~\ref{sec:interaction_and_human_learning}. Key implementation parameters are summarized in
Table~\ref{tab:forcerft_parameters} in the Appendix.
\subsection{Performance on Contact-Rich Tasks}
\label{sec:comparative_results}

SmolVLA-SFT struggles with contact-dependent recovery:
once an initial insertion or ring-placement attempt fails,
subsequent motions rarely complete the task.
The force-conditioned prior enables gradual insertion
corrections after such failures, providing more
contact-responsive base motions. However, it does not
learn from deployment outcomes, and both offline baselines
still exhibit missed ring grasps.
Residual imitation further improves completion through
human corrective supervision, but insertion and assembly
failures remain dominated by timeouts, with premature
gripper release also observed before full insertion.
ForceRFT additionally uses task rewards to refine these
corrections, improving insertion and assembly from
19/30 and 22/30 for residual imitation to 25/30 and 27/30,
respectively (Table~\ref{tab:main_results}).
Its five insertion failures are all timeouts; the three
assembly failures comprise two timeouts and one ring drop
following a substantial contact-induced change in the
ring's pose within the gripper.
Wiping improves progressively across the four methods,
but the gain over residual imitation is smaller
(8/30 to 11/30). A single pass often leaves marks,
requiring repeated cleaning followed by lifting the
eraser and stopping. Both incomplete cleaning and
continued wiping after text removal persist across methods.
Together, these results support combining force-conditioned
motion generation with value-guided residual correction,
while showing that the gains on insertion and assembly
do not yet extend to equally reliable cleaning and
autonomous termination.
\begin{table}[t]
    \centering
\caption{Real-robot performance.
Autonomous successes out of 30 trials for each task.}
    \label{tab:main_results}
    \small
    \setlength{\tabcolsep}{3pt}
    \renewcommand{\arraystretch}{1.08}
    \begin{tabularx}{\columnwidth}
        {l*{3}{>{\centering\arraybackslash}X}}
        \toprule
        Method
        & \shortstack{Plug\\insertion $\uparrow$}
        & \shortstack{Ring-on-peg\\assembly $\uparrow$}
        & \shortstack{Whiteboard\\wiping $\uparrow$} \\
        \midrule
        SmolVLA-SFT
        & 7/30 & 10/30 & 2/30 \\
        Force-conditioned prior
        & 12/30 & 17/30 & 4/30 \\
        Residual imitation
        & 19/30 & 22/30 & 8/30 \\
        \midrule
        \rowcolor{gray!12}
        \textbf{ForceRFT}
        & \textbf{25/30}
        & \textbf{27/30}
        & \textbf{11/30} \\
        \bottomrule
    \end{tabularx}
\end{table}
\begin{table}[t]
    \centering
\caption{Wrist-feedback ablation on plug insertion.
Autonomous successes out of 30 trials.}
    \label{tab:wrist_ablation}
    \small
    \setlength{\tabcolsep}{4pt}
    \renewcommand{\arraystretch}{1.08}
    \begin{tabularx}{\columnwidth}
        {>{\raggedright\arraybackslash}Xc}
        \toprule
        Configuration & Plug insertion $\uparrow$ \\
        \midrule
        w/o wrench increment & 23/30 \\
        w/o residual wrist feedback & 19/30 \\
        \midrule
        \rowcolor{gray!12}
        \textbf{Full (ForceRFT)} & \textbf{25/30} \\
        \bottomrule
    \end{tabularx}
\end{table}
\subsection{Ablation Studies}
\label{sec:wrist_feedback_ablation}

\noindent\textbf{Wrist-Feedback Ablation.}
All variants retain the same frozen force-conditioned
prior but differ in the wrist inputs available to
the residual actor and critics (Table II).
Removing both $\mathbf w_n$ and $\Delta\mathbf w_n$ produces the larger
degradation, reducing insertion success from 25/30
to 19/30. This suggests that force conditioning during
base-motion generation does not replace direct feedback
at execution: the residual learner still benefits from
contact information acquired after the action chunk
is generated. Removing only $\Delta\mathbf w_n$ retains 23/30,
indicating that the current wrench preserves most of
the observed benefit. The smaller difference provides
more limited evidence for the additional contribution
of explicit wrench dynamics.

\noindent\textbf{Online Adaptation Ablation.}
Residual imitation removes the value objective while
retaining wrist conditioning, human corrective supervision,
and residual regularization.
Its lower final performance across all three tasks
(Table~\ref{tab:main_results}) supports return-guided
refinement beyond matching local recovery actions:
corrective imitation supplies action targets, whereas
value learning additionally evaluates corrections
through their autonomous outcomes.

\begin{figure}[t]
    \centering
    \includegraphics[width=0.9\columnwidth]
        {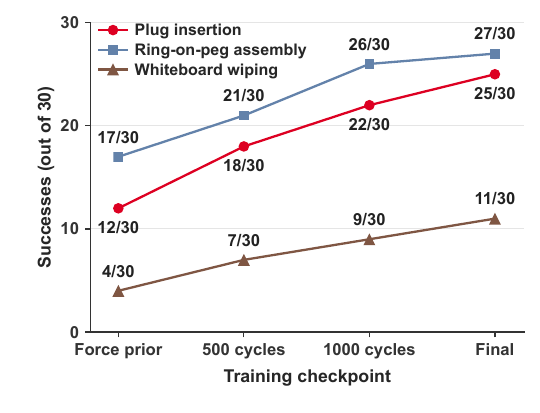}
    \caption{\textbf{Online adaptation across three tasks.}
    Successes out of 30 autonomous trials per task for
    the frozen prior and ForceRFT checkpoints.
    Connecting lines indicate checkpoint order,
    not equal cycle intervals.}
    \label{fig:insertion_checkpoints}
\end{figure}

Fig.~\ref{fig:insertion_checkpoints} complements this
comparison by examining the full model during training
across all three tasks.
We evaluate at 500-cycle intervals and at the budget
endpoint to retain intermediate measurements within
the 30-minute online budget.
Under the data-dependent schedule in
Section~\ref{sec:interaction_and_human_learning},
eight unique TD-eligible autonomous transitions grant
one joint cycle; a 500-cycle block therefore consumes
the allowance supplied by 4000 such transitions,
rather than permitting unrestricted replay updates.
Across all three tasks, success counts exceed those
of the frozen prior at 500 cycles and increase at
each subsequent sampled checkpoint.
Together, the input and value-objective ablations
support both execution-time contact conditioning
and reward-guided residual refinement.

\section{CONCLUSION}
\label{sec:conclusion}
We presented ForceRFT, a force-guided residual reinforcement
learning framework that combines a frozen force-conditioned
VLA prior with online command-level adaptation.
Execution-time wrist wrench and its temporal change condition
both residual correction and twin-critic value estimation.
The residual policy learns from autonomous task outcomes
and human corrective supervision while the prior remains fixed.
Real-robot experiments show higher autonomous success rates
than the evaluated demonstration-trained and
residual-imitation baselines on plug insertion,
ring-on-peg assembly, and whiteboard wiping.
Comparisons with residual imitation support value-guided
refinement, while plug-insertion ablations indicate the
benefit of direct wrist feedback.
Checkpoint evaluations further show increasing autonomous
success across the sampled training stages in all three tasks.
Future work will explore how to coordinate local contact
recovery with task-level progress in multi-stage manipulation.
% Summarize the findings without repeating the abstract verbatim.
% Briefly state limitations and future directions when space permits.

% Do not include acknowledgments in the anonymous submission if they may
% reveal author identities. Add them back only in the camera-ready version.

\bibliographystyle{IEEEtran}
\bibliography{references}
\appendix[Implementation Details]
\label{app:implementation}

Table~\ref{tab:forcerft_parameters} summarizes the key
settings. The residual actor and each critic use two
256-unit hidden layers with SiLU activations.
SFT uses 500 linear warm-up updates followed by
cosine decay to $2.5\times10^{-6}$.

\begin{table}[H]
    \centering
    \caption{Key Implementation Parameters}
    \label{tab:forcerft_parameters}
    \footnotesize
    \setlength{\tabcolsep}{5pt}
    \renewcommand{\arraystretch}{1.15}

    \begin{tabularx}{\columnwidth}{
        @{}
        >{\raggedright\arraybackslash}X
        >{\raggedright\arraybackslash}p{0.28\columnwidth}
        @{}
    }
        \toprule
        Parameter & Value \\
        \midrule

        SFT updates
        & $10{,}000$ \\

        SFT effective batch size
        & $4$ \\

        SFT optimizer
        & AdamW \\

        SFT peak learning rate
        & $10^{-4}$ \\

        Euler integration steps
        & $10$ \\

        \midrule

        Online optimizer
        & Adam \\

        Actor learning rate
        & $3\times10^{-5}$ \\

        Critic learning rate
        & $3\times10^{-4}$ \\

        TD batch size
        & $128$ \\

        Actor-Q batch size
        & $64$ \\

        Human imitation batch size
        & $32$ \\

        Discount factor $\gamma$
        & $0.99$ \\

        Target mixing rate $\rho$
        & $0.005$ \\

        Value loss weight $\lambda_Q$
        & $0.1$ \\

        Imitation loss weight $\lambda_H$
        & $1.0$ \\

        Residual penalty weight $\lambda_R$
        & $0.1$ \\

        \midrule

        Normalized residual cap $c_{\mathrm{norm}}$
        & $0.1$ \\

        Translation residual limit per axis
        & $1\,\mathrm{mm}$ \\
        
        RPY residual limit per axis
        & $0.5^\circ$ \\

        Nominal decision rate
        & $10\,\mathrm{Hz}$ \\

        Wrench low-pass cutoff
        & $8\,\mathrm{Hz}$ \\

        \bottomrule
    \end{tabularx}

    \par\smallskip
    \begin{minipage}{\columnwidth}
        \footnotesize
        \raggedright
        The shared residual bound is
        $c_{\mathrm{res}}
        =\min\{c_{\mathrm{norm}},
        \min_j d_j^{\max}/\sigma_j\}$,
        where $d_j^{\max}$ is the per-axis residual limit
        listed above and $\sigma_j$ is the corresponding
        frozen SFT action standard deviation.
        Both use matching units, with RPY in radians.
    \end{minipage}
\end{table}

% Keep your existing bibliography commands below.

\end{document}